# Application of deep learning approaches for medieval historical documents transcription*

Maksym Voloshchuk[1, †], Bohdana Zarembovska[1, †], Mykola Kozlenko[2,*, †]

[1] *Vasyl Stefanyk Carpathian National University, Shevchenka 57, 76018 Ivano-Frankivsk, Ukraine*
[2] *SoftServe Inc, 201 W 5th Street, Suite 1550, Austin, TX 78701, USA*

### Abstract
Handwritten text recognition and optical character recognition solutions show excellent results with processing data of modern era, but efficiency drops with Latin documents of medieval times. This paper presents a deep learning method to extract text information from handwritten Latin-language documents of the 9th to 11th centuries. The approach takes into account the properties inherent in medieval documents. The paper provides a brief introduction to the field of historical document transcription, a first-sight analysis of the raw data, and the related works and studies. The paper presents the steps of dataset development for further training of the models. The explanatory data analysis of the processed data is provided as well. The paper explains the pipeline of deep learning models to extract text information from the document images, from detecting objects to word recognition using classification models and embedding word images. The paper reports the following results: recall, precision, F1 score, intersection over union, confusion matrix, and mean string distance. The plots of the metrics are also included. The implementation is published on the GitHub repository.

## 1. Introduction

Before Gutenberg invented the printing press in 1439, all texts were handwritten. This aspect complicates the process of reading such documents because of the variety of fonts and approaches to writing text by hand. Some handwritten fonts require a scientist to have a lot of experience to understand a document's payload.

### 1.1. First-sight overview of the given historical documents

In the case of medieval documents, we must keep in mind the age of such texts, which might be more than a thousand years. Because of that, the state of some documents might have some impairments, for example, lost fragments, stains over the text, etc. The examples of such deficiencies are shown on Figure 1.

Given from the Center of Medieval Studies, Latin-language act documents are from the chancellery of the Carolingian and Ottonian dynasties of the 9$^{th}$-11$^{th}$ centuries, which are publicly available at [1, 2]. The material of the documents mostly describes the history of the formation of Latin church dioceses. Most of the documents are created on a parchment base and have a relatively well-preserved state. The text is written in Carolingian minuscule script with elements of Carolingian majuscule, 8$^{th}$-9$^{th}$ century. An example of such a document is depicted in Figure 2. We can see that the content is divided by text lines, and the font may vary within the same document.

[1*] Corresponding author.
[†] These authors contributed equally.
maksym.voloshchuk.21@pnu.edu.ua (M. Voloshchuk); bohdana.zarembovska.21@pnu.edu.ua (B. Zarembovska); mykola.kozlenko@cnu.edu.ua (M. Kozlenko)
0009-0005-4950-6234 (M. Voloshchuk); 0009-0002-7673-5970 (B. Zarembovska); 0000-0002-2502-2447 (M. Kozlenko)

We can see some structure in the depicted example. The first line of the document almost always has a distinct font, also it’s provided with the first capital letter "C", written in a decorative style. The beginning of a document is almost always the same: "In nomine sanctae et individuae trinitatis". The main payload of the document is written after the first line – it describes the main ideas and events. After the main middle part, there is a section for the initials, decorative elements, and the seal. Almost every text ends with the word "Amen", including the one in the example.

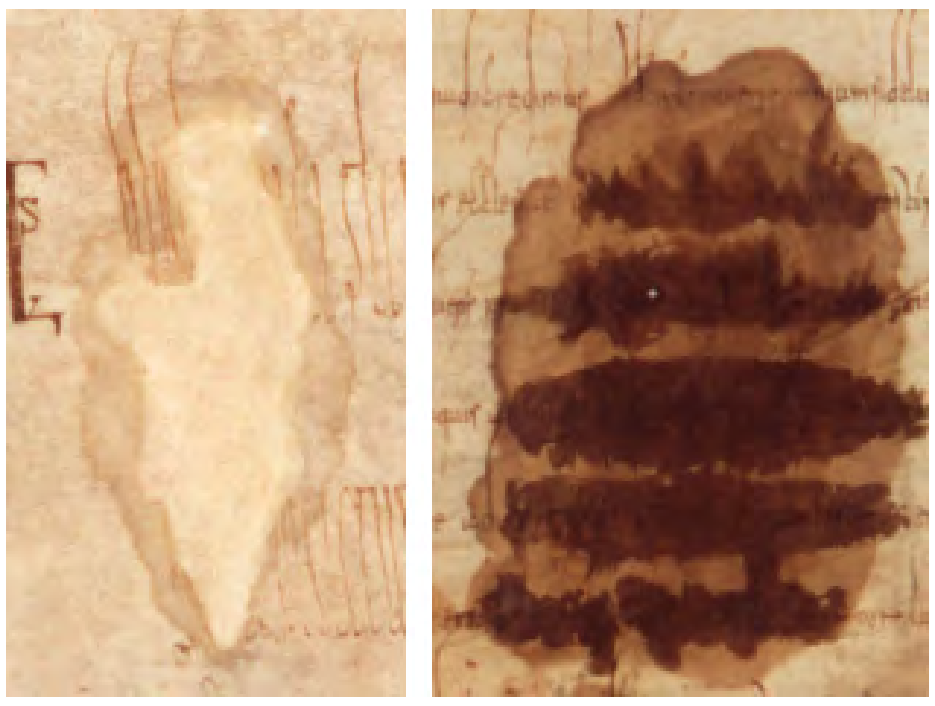

**Figure 1**: Example of medieval documents' deficiencies.

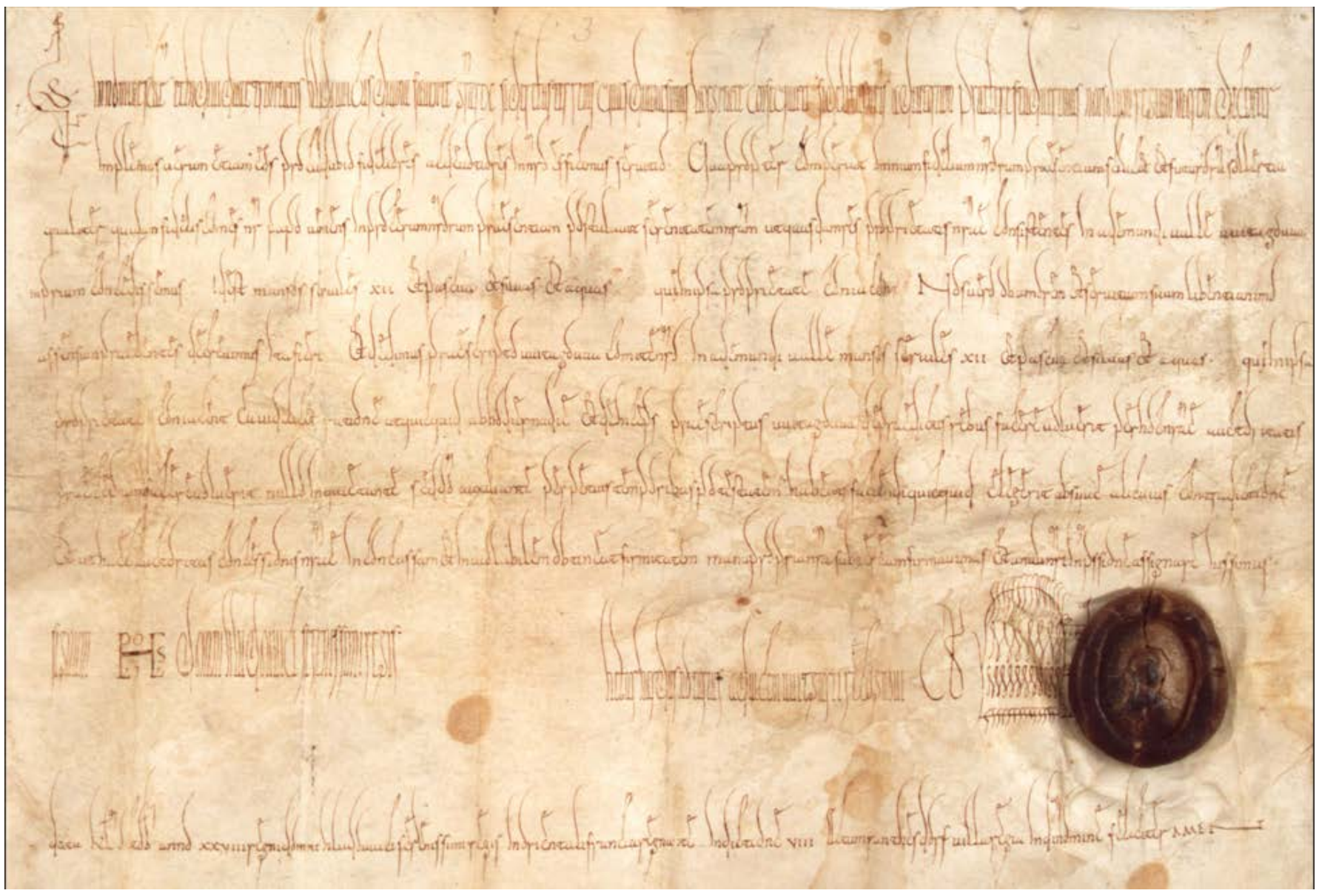

**Figure 2**: Example of the document from the given database.

You do not need to be a historian to distinguish some of the words since they are visually clearly separated. But often in the text, it is not so clear that the two parts of a writing are separate words, for example, "In nomine" at the beginning of the documents is often written as a single word without any separator. Also, there is a system of word contraction, so some of the words are not written as they are spelled, but are present in a shortened version. If you’re studying such documents as a historian, you need to know all the details about the scripting culture of the time and place where the document was written.

The given database includes 31 images of documents and corresponding texts previously extracted by historians. The image quality is enough to recognize words and conveniently read the text of the document. The transcribed texts, except for everything else, contain the content of damaged parts and the full variation of shortened words. Also, the metadata about the document is included, for example, date, special notes, etc.

Some of the documents aren’t the original ones; they are copies made in later centuries. The overall style and font of such images are noticeably different, and their preservation state is much better.

### 1.2. Related work

Handwritten Text Recognition (HTR) approaches have already done a great job in transcribing handwritten texts. There are a lot of studies on applying HTR methods to recognize text on historical documents. The transformer-based models have already shown accurate results in the case of historical HTR [3]. The authors have used 16th-century Latin texts containing around 17500 text lines to train a transformer-based model to transcribe the texts.

The most common approach in HTR is to extract text lines and recognize text inside the lines using models trained by connectionist temporal classification (CTC) loss. The models that use CTC loss originally were built for speech recognition problems, but similar approaches are effective in the case of recognizing handwritten symbols. The OrigamiNet [4] proposes a solution able to extract text not only from text lines but also from whole pages. The model architecture consists of a CNN backbone and the CTC head part. The bidirectional LSTM module might be used to make the model remember previous states. Also, the Attention mechanism and language models can be included, which is shown in [5].

The extraction of text lines is preceded by semantic segmentation in most cases [6]. The most common approach to achieve the goal is to build a U-Net model [7]. The application of the segmentation model is useful in cases of clearly distinct text lines, even if their shape is curvy, as in the case of [6], which has shown good results in dealing with such scenarios.

In the field of object detection, the YOLO architecture shows state-of-the-art results [8]. The large variety of YOLO models is widely used in real-world scenarios, including inference on edge devices such as Raspberry Pi boards. Nowadays, the YOLO models can solve not only the detection problem, but also visual object tracking, semantic segmentation, and pose estimation. The Ultralytics YOLO implementation allows for easy training and deployment/export of detection models. The efficiency of using YOLO in visual elements detection of different classes was shown in [9]. Nowadays, YOLO capabilities are not constrained only by object detection; there are versions for segmentation, tracking, and pose estimation problems [10]. Segmentation and detection models can also be used for document Layout analysis [11]. Paper [12] demonstrates the application of deep learning for the detection problem. Paper [13] introduces semantic segmentation using deep learning and artificial neural networks.

## 2. Problems to be solved

The majority of HTR and optical character recognition (OCR) models are trained on modern data, whose age is no more than 300 years. In most cases, the handwritten texts are well preserved and written in languages that are currently used today and have not changed very much. If medieval data is fed to the models, the processing results are not accurate enough.

The second problem concerns CTC models – these models do not provide information about the position of extracted words, which is useful for analysis. For historians as end users, it’s hard to know what exact information is used by the model to make the prediction of a specific word. OCR systems can solve this task for modern data, but it is not effective with medieval text. Also, the medieval Latin texts often contain a significant number of contracted words, which makes it harder to transcribe a text character by character. For example, “nostrae” (ours in English) often is written using only 3 characters – “nse”.

To train a word classification model to make accurate predictions, there is a requirement to cover a major part of the vocabulary. It causes a problem if there is a need to work with a limited amount of data.

This paper proposes an approach to solve these problems in a single modular architecture solution. It uses visual object detection models to extract objects’ coordinates and a pair of CNN models to interpret the detected words.

# 3. Data

## 3.1. Development of the training datasets

We've annotated our document images with words and text lines. The word coordinates are defined as a bounding box: *(x, y, w, h)*, where *x, y* – are the coordinates of the box's center; *w, h* – width and height, respectively. An example of a labeled document is depicted in Figure 3. We decided to label the document using different classes for more accurate further use.

The ordinal bounding box is not a proper format for text line annotation because of its strict rectangular shape, whose sides are parallel to the image sides. Often in handwritten documents, not only medieval ones, you might encounter that the text lines are not perfectly aligned, so another format of annotation is needed. The solution was the oriented bounding boxes – the main difference is that it's defined by 8 numbers – $(x_1, y_1, x_2, y_2, x_3\ y_4)$, where each $(x_i, y_i)$ pair is coordinates of the corresponding corner. The given format allows to make more flexible annotations, which are needed for extracting text lines from the document image. Also, we can deduce the angle at which the text line is inclined.

To train a model to detect word instances inside cropped text lines, there is a need to create a separate dataset that contains images of text lines. Our team wrote a Python script to define which word belongs to which line using the given formula (1) as a criterion.

$$\frac{|w \cap l|}{|w|}, \qquad (1)$$

where $w$ – bounding box of a word, and $l$ – bounding box of a text line, the $|.|$ operation represents an area of a 2D shape. The 0.5 threshold value was sufficient for this task. The visualization of the process of defining element relationships is shown in Figure 4.

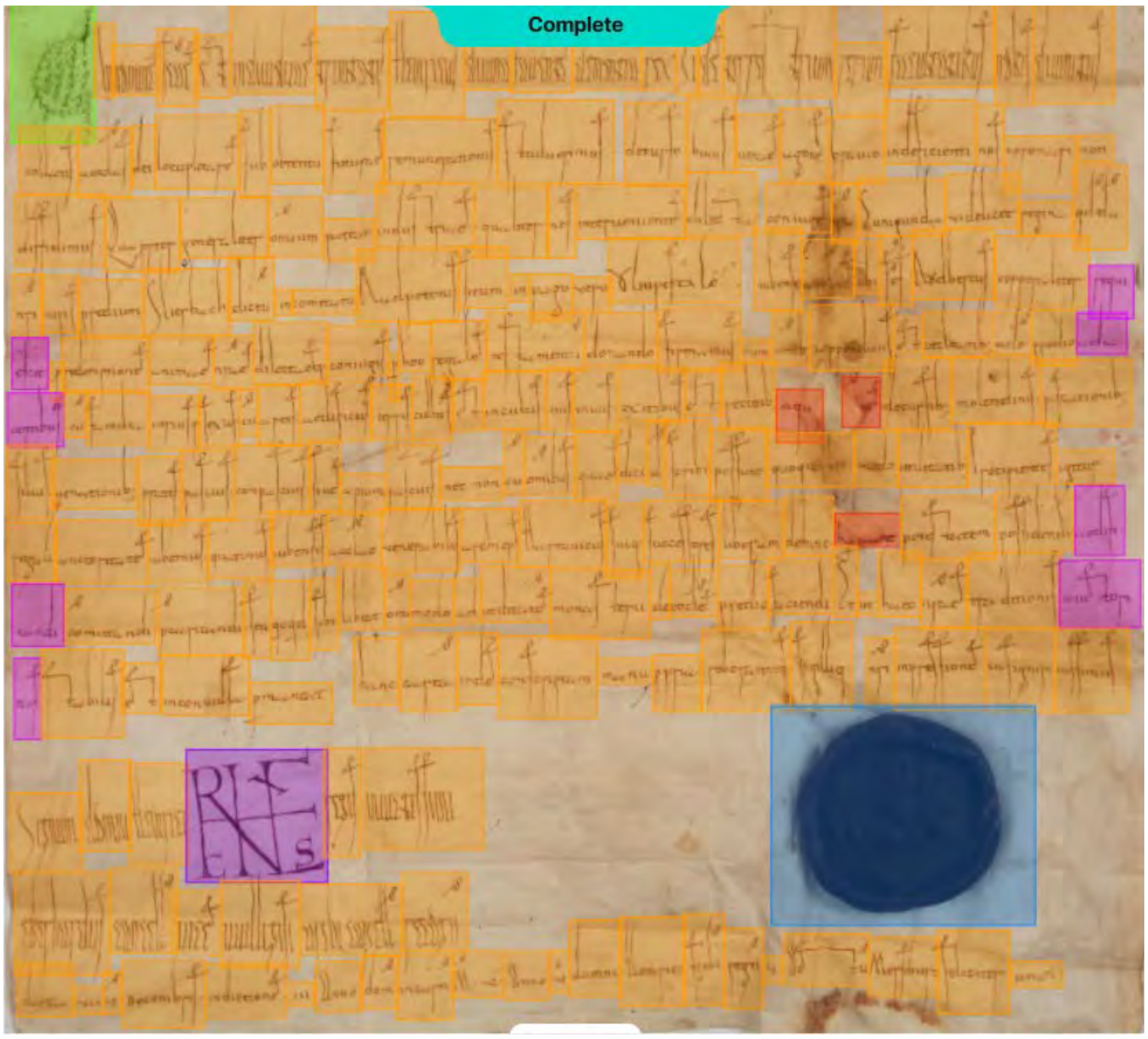

**Figure 3**: Medieval document with labeled elements.

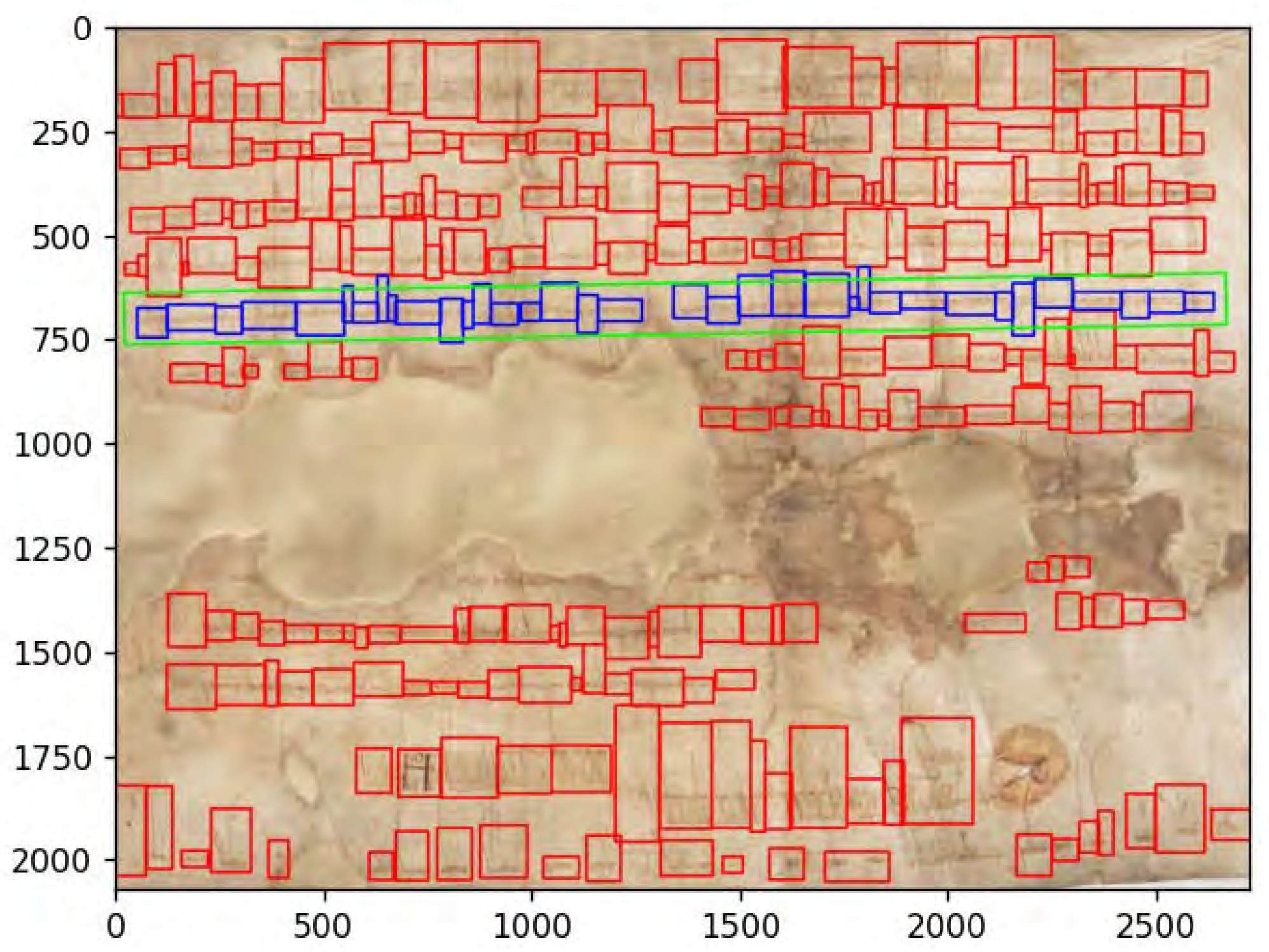

**Figure 4**: Searching for words that belong to a line. Green shape – bounding box of a line, blue shapes – bounding boxes of words belonging to the line, and red shapes – bounding boxes of words not belonging to the line.

To make each text line aligned to the image sides, we need to apply rotation to the cropped text line image. The center of the cropped image is the center of rotation. Let's consider the top and bottom sides as vectors pointing right. The text line direction vector is defined as the mean value of the top and bottom vectors. To calculate the angle between the line direction vector and a horizontal line, the (2) is used.

$$\alpha = \arctan\left(\frac{y_t + y_b}{x_t + x_b}\right), \quad (2)$$

where $x_t$, $y_t$ – coordinates of top direction vector, $x_b$, $y_b$ – coordinates of bottom direction vector. To make data consistent, we need to rotate the words' coordinates as well. To reach it, the rotation matrix with an inverted angle is applied for each word center point, and the width and height are left with the same values.

For word classification, we need a separate dataset to map word images to their text representation. We matched labeled word annotations to words in transcribed text. Corrupted words and words with carry to the next line are not included in the classification dataset.

## 3.2. Explanatory Data Analysis

In total, the dataset contains 31 documents. Mean width and height of the document images are 2733 and 2246 pixels, respectively. Each document contains 300 words on average, but there are several outliers containing more than 1200 words. The distribution of word numbers across the documents is shown in Figure 5. The dataset contains 610 text lines in total and 19.67 lines per document on average. There are 4800 annotated word objects in the whole dataset. Each text line contains 20.65 words on average.

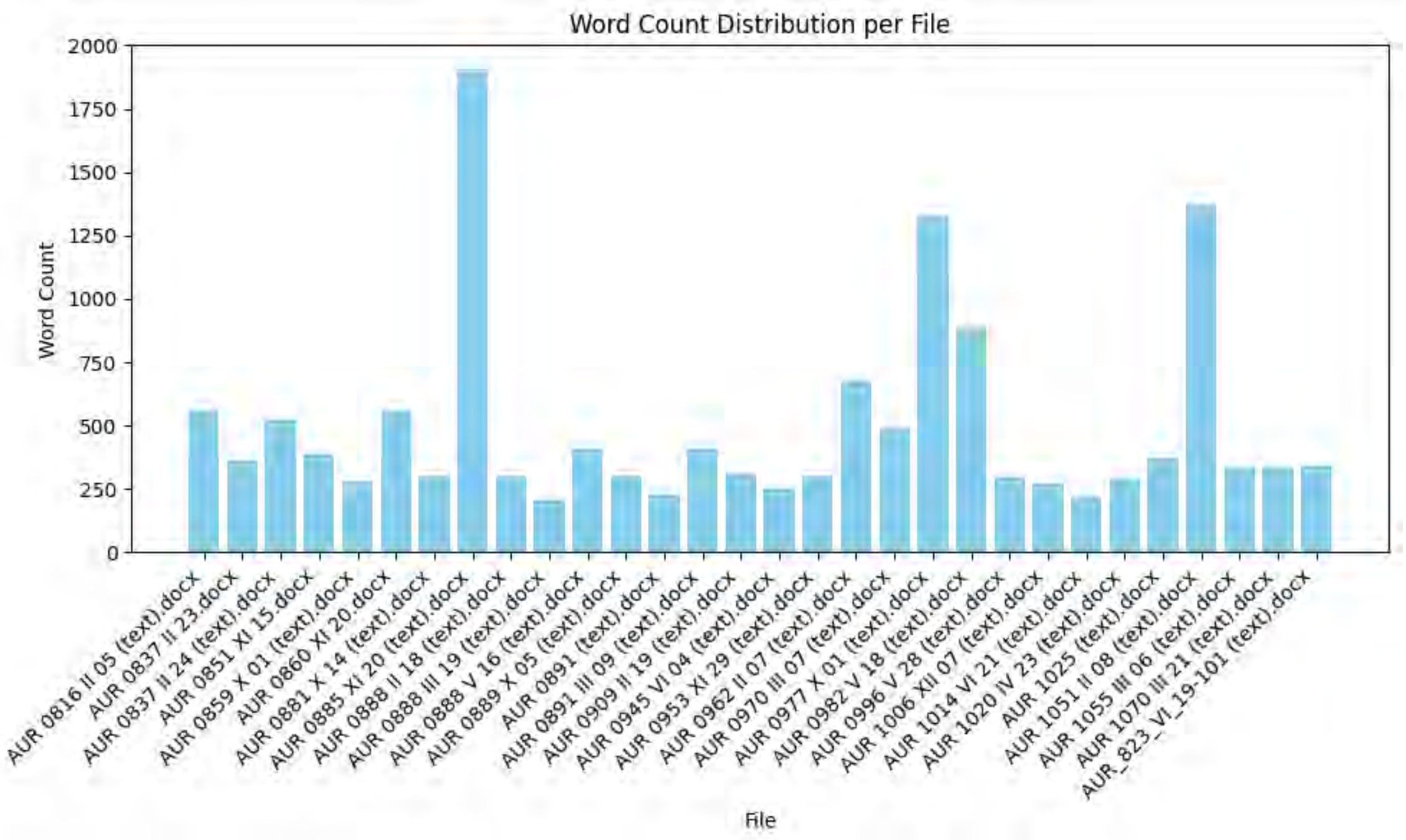

**Figure 5**: Word number distribution across the documents.

Conjunctions are the most common words, such as "et", "in", "ad", "cum", and so on. 2433 words occur 1 or 2 times in the whole dataset. The whole distribution of word occurrences is depicted in Figure 6. 2433 words occur 1-2 times, 1012 words occur 2-5 times, 289 words occur 5-10 times, 206 words occur 10-25 times, 56 words occur 25-50 times, 17 words occur 50-100 times, 6 words occur 100-500 times.

Latin has an extensive declension system, so cognate words in different cases are syntactically similar. To measure the similarity between two words modified Hamming distance is used [13].

We propose a modified Hamming distance calculation that incorporates length-based constraints to handle minor structural differences between words. If the absolute length difference exceeds 2, the function assigns an infinite distance, indicating excessive dissimilarity. For words of identical length, the traditional Hamming distance is computed as the count of differing characters. If the lengths differ by one, the function checks whether removing a single character from the longer word results in a match, assigning a distance of 1 if successful. Additionally, words shorter than six characters are deemed incomparable to words of six or more characters, further refining the comparison criteria.

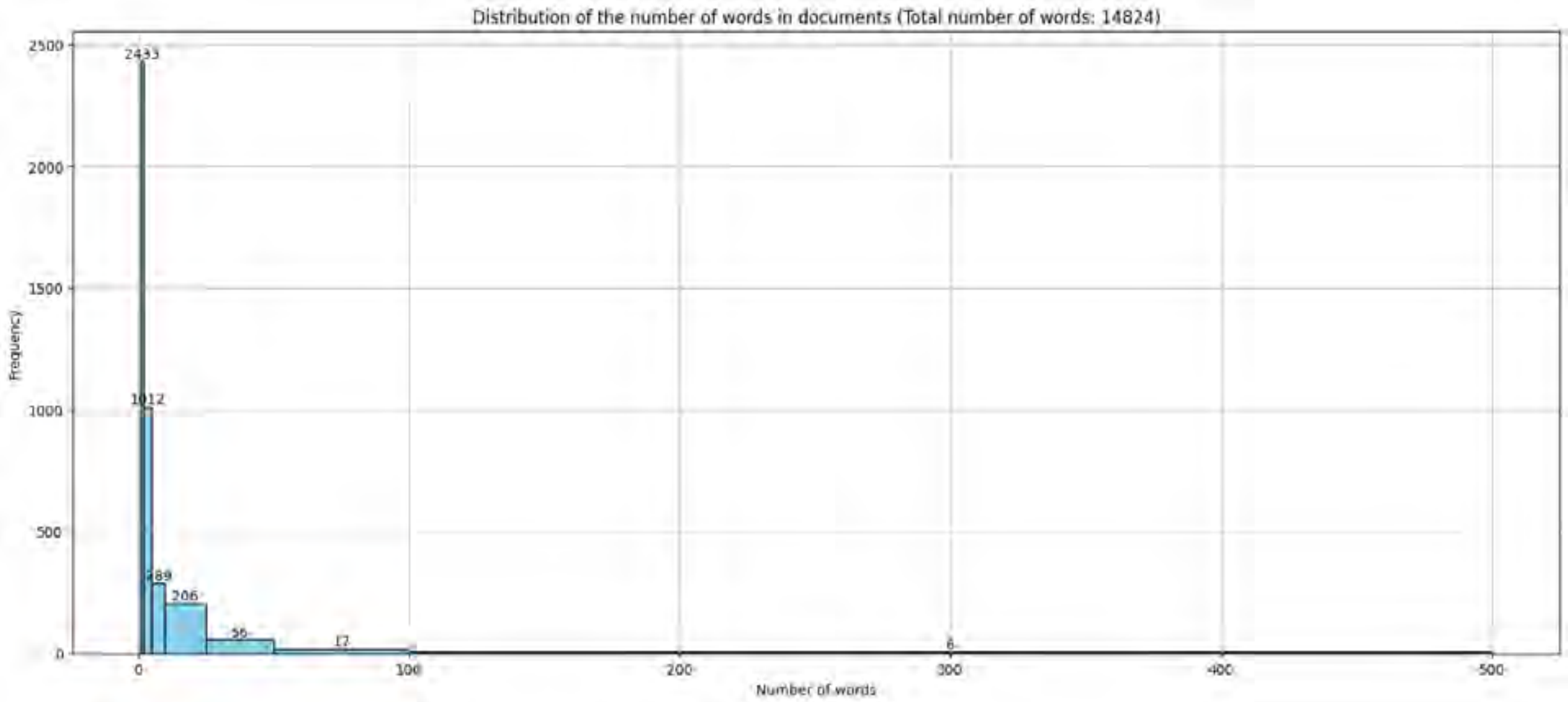

**Figure 6**: Distribution of word occurrences.

The total number of similar word pairs is 2406. After merging the similar words (considering them as the same ones), we encounter that the number of rare words is significantly reduced: only 1868 words occur 1-2 times, and 989 words occur 2-5 times.

# 4. Deep learning approaches

To reach the goal of extracting text from medieval documents, we developed a pipeline using detection and classification approaches. Also, we provided a model to map the word images onto a linear space to make it possible to find similar words using images only.

To do so, we use several deep learning techniques, which have shown their efficiency in a large number of domains. Most commonly, in our research, we used the transfer learning technique [15], which is commonly used in computer vision.

Surely, not all the transferred weights are relevant, so we might need to train the model with transferred information. Such a training process, in most cases, is much quicker than training a model from scratch – transferred weights make the model's loss value much closer to the optimum point at the start of the training.

Also, we used the fine-tuning technique to train our detection model to deal with more specific data [16].

## 4.1. YOLO for text line detection

YOLO is a widely used detection architecture that can make fast detections using a single model pass. The Ultralytics implementation of these models makes it easy to use detectors in different projects.

The line detector has been trained for 300 epochs and reached a result of 98% precision and 93% recall. To train the model, the rotation augmentation was used, and the rotation angle was in the range from -5° to 5°. The scaling and translation were also applied for the augmentation.

The Adam optimization algorithm was used with a scheduled learning rate: $8 * 10^{-4}$ as a starting value and $10^{-6}$ as a finish value. The batch size value was set to 6. Also, the lost function weights have to be set for the most effective optimization. Since we predict a single class on the image, the classification loss is not that important, so its weight value was reduced to 1.5. The box loss weight was raised to 9, and the distribution focal loss was set to 1.3.

The training images were of a square shape of size 700 pixels. The model was trained on RTX 3050 mobile GPU with 4 GB of video memory.

Some of the line predictions might intersect each other, so we had to resolve such scenarios as a post-processing step. We've tried several methods to solve this problem. To detect intersections, the intersection over union (IoU) was used (3).

$$\mathrm{IoU}(\mathrm{a},\mathrm{b}) = \frac{|a \cap b|}{|a \cup b|}, \tag{3}$$

where $a$ and $b$ are bounding boxes, and the $|.|$ operator represents the area of a shape.

We've chosen to use 0.4 as a threshold value and to IoU of two bounding boxes, and we've defined the intersection of two shapes if the IoU value is higher than or equal to the threshold.

We used confidence value to resolve intersections for text line detection, meaning if two bounding boxes are intersecting, leave one with a higher confidence value.

After all the intersections are resolved, we need to extend the line bounding boxes to the image's right and left corners while maintaining the angle of inclination. Our extending method moves the top and bottom bounding box edges by calculating their corner coordinates. The horizontal coordinates became 0 or 1 depending on whether it's the left or right side of the bounding box, and the new vertical coordinates are calculated using an average of the top and bottom vectors.

After the bounding box extension is done, the lines are cropped and rotated to equalize the text they contain. To get the rotation angle, the average inclination of the bounding box's top and bottom corners is used. The result of line detection with post-processing is depicted in Figure 7.

After the post-processing, the cropped lines are used to detect words inside. The cropped lines are sorted from top to bottom to maintain the right text order.

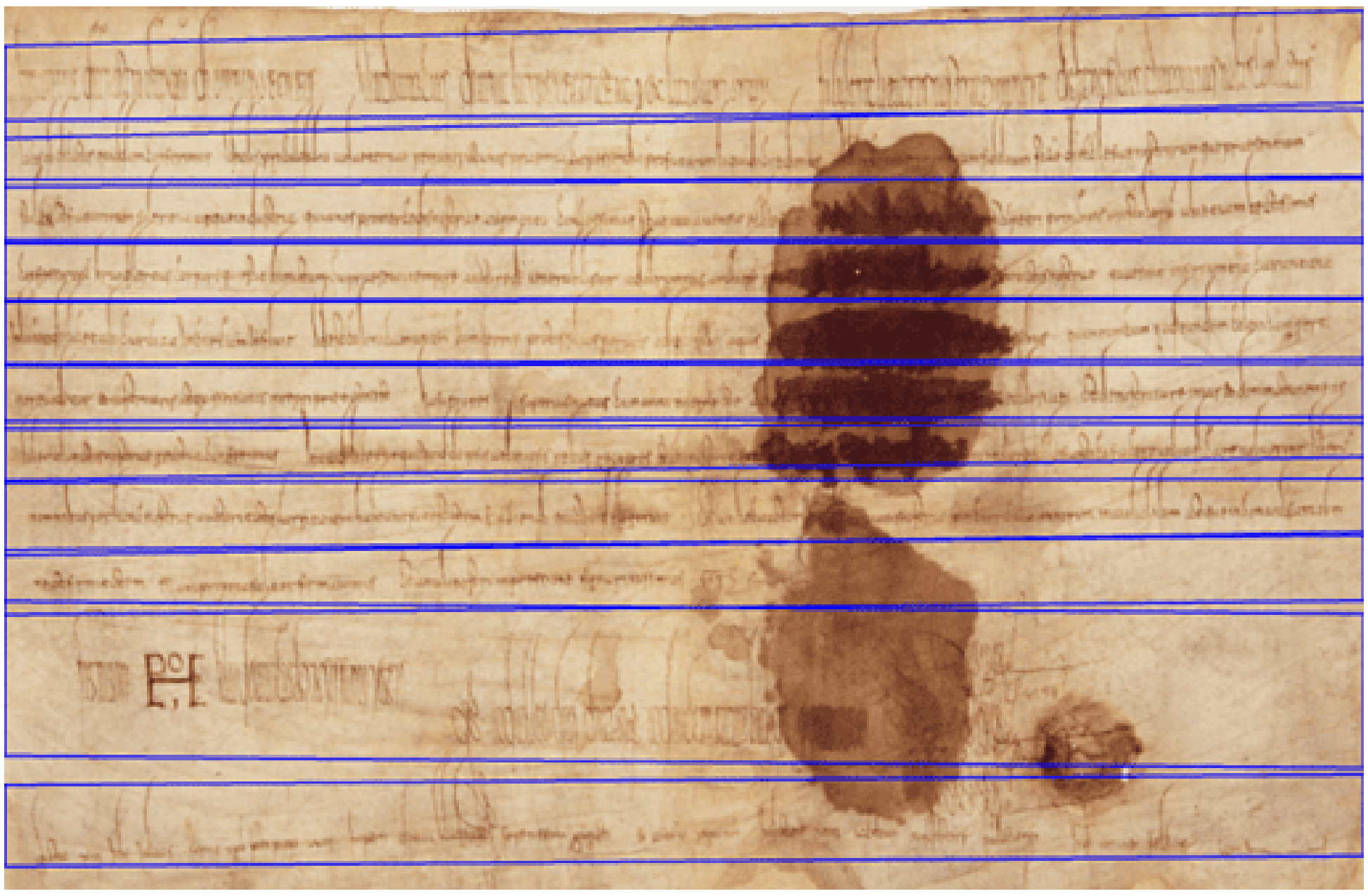

**Figure 7**: Result of detecting text lines on an “unseen” by the model historical document.

## 4.2. YOLO for word detection

For word detection inside the text lines, we fine-tuned the "yolov8m" model (8th medium YOLO version). The hyperparameter tuning was the most valuable process to train the model. The fine-tuned model with default hyperparameter values resulted in too low recall to use it.

We trained our detection model for 400 epochs. Most of the hyperparameters are copied from the training of text line detection, but with some significant changes. Because there was only 4 GB of video memory available, the batch size value was set to 4. The image size was 1024 pixels in a rectangular shape to maintain the aspect ratio. The augmentation was also changed; only the translation and cropping were applied for the training process. The example of a training batch is depicted in Figure 8.

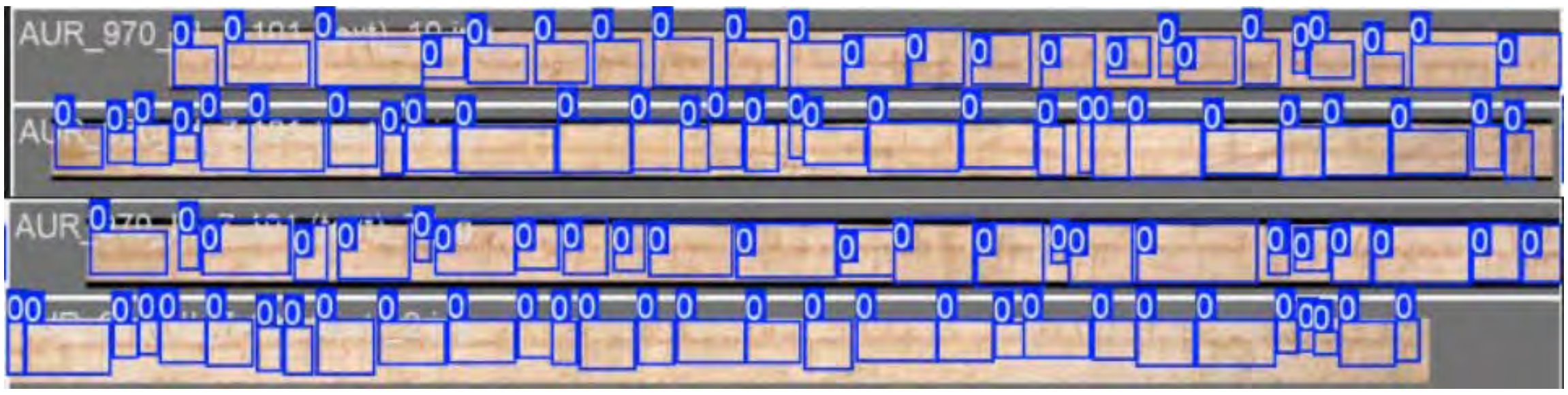

**Figure 8**: Training batch of the word detection model.

We used a confusion matrix, F1-Confidence curve, Precision-Confidence curve, Recall-Confidence curve, and Precision-Recall curve to compare our trained models and choose the best one for further use. To deal with the intersections, the union resolving has shown the most effective results, meaning if two bounding boxes are intersecting too much, return the smallest bounding box, which contains both of the intersecting ones. The trained model’s results of detecting words inside a cropped text line are shown in Figure 9.

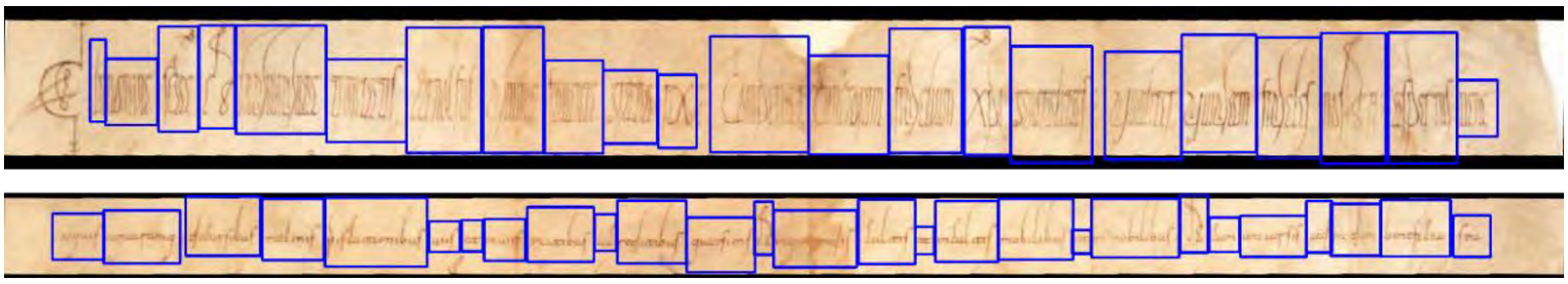

**Figure 9**: Word detection results. Union intersection resolving is applied as a post-processing step.

After the prediction is done, the words are cropped from text line images and sorted from left to right to maintain text order. Then the words are passed to the classification model.

### 4.3. Word classification model

The decision to choose an approach with word classification instead of processing text by characters was motivated by the large number of abbreviated words in the medieval documents. For example, the word "nostrae" is often contracted into 3 letters.

We have built our classifier by combining a pre-trained ResNet50 backbone and an MLP head network. The head neural network consists of a dropout layer and three linear layers of size 2048, which are separated by a ReLU activation function and the Softmax activation at the end of the network. The task of the model is to classify the word written on an input image. Compared to CTC models, the classification architecture is not affected by abbreviated words and the variety of ways to write characters.

The training input images are augmented using scaling, rotations, and elastic transformation, and resized to a shape of 200x200 pixels. Our dataset represents the distribution of words in real human language, so some words occur with much higher frequency than others; in other words, the dataset is unbalanced. Augmentation is a crucial part of dealing with highly unbalanced data, but the other step is applying learning weights for each class. The weights are calculated as the inverse of word occurrences in the dataset, so the word occurring once has the weight 1, the word occurring twice has the weight 0.5, and so on. The train/validation split was performed in such a way as to ensure that the words, which occur only once, are included in the train subset, so that while training the model encounters representations of all classes. The output size of the model is 1765, which is the number of known words in our dataset. This value will increase with the size of the training data.

### 4.4. Word similarity measure model

The words classification dataset is highly unbalanced, so besides classification, we need some additional tools to recognize words. Our proposed method is to build an embedding model, which maps a given word image to a linear space such that the vector representation of images with similar words are the closest ones in terms of Euclidean distance. The method is based on the FaceNet approach [10].

We use a pretrained ResNet50 CNN model as a backbone to extract visual features. The extracted features are then processed by a network consisting of 3 linear layers with ReLU activation functions and a single residual connection. The model architecture is shown in Figure 10.

We used triplet margin loss as a criterion function and Adam optimizer with $5 * 10^{-5}$ learning rate value. The model was trained for 400 epochs with 400 training triplets for each epoch and with a batch size value of 16. The image size was set to 120 since it's half of mean size of the cropped word images. The margin value for the triplet loss function was set to be 2. The output size of the model was set to be 64.

The words, whose string distance is less than or equal to 1 (same words or differ at 1 symbol), are defined as similar. So, the train triplets are formed in such a way that the anchor and positive images represent similar words, and the anchor and negative images represent different words.

After the training was done, all the training images were mapped onto the embedding space using the trained model. The part of the TSNE 2d projection of the embeddings is depicted in Figure 11. For inference, the model receives an unseen image of a word and outputs its embedding representation; to find the most similar words, we need to find the closest embedding vectors in the created space. To make the closest embedding search faster than just linear time brute force, the Faiss vector database is used [11].

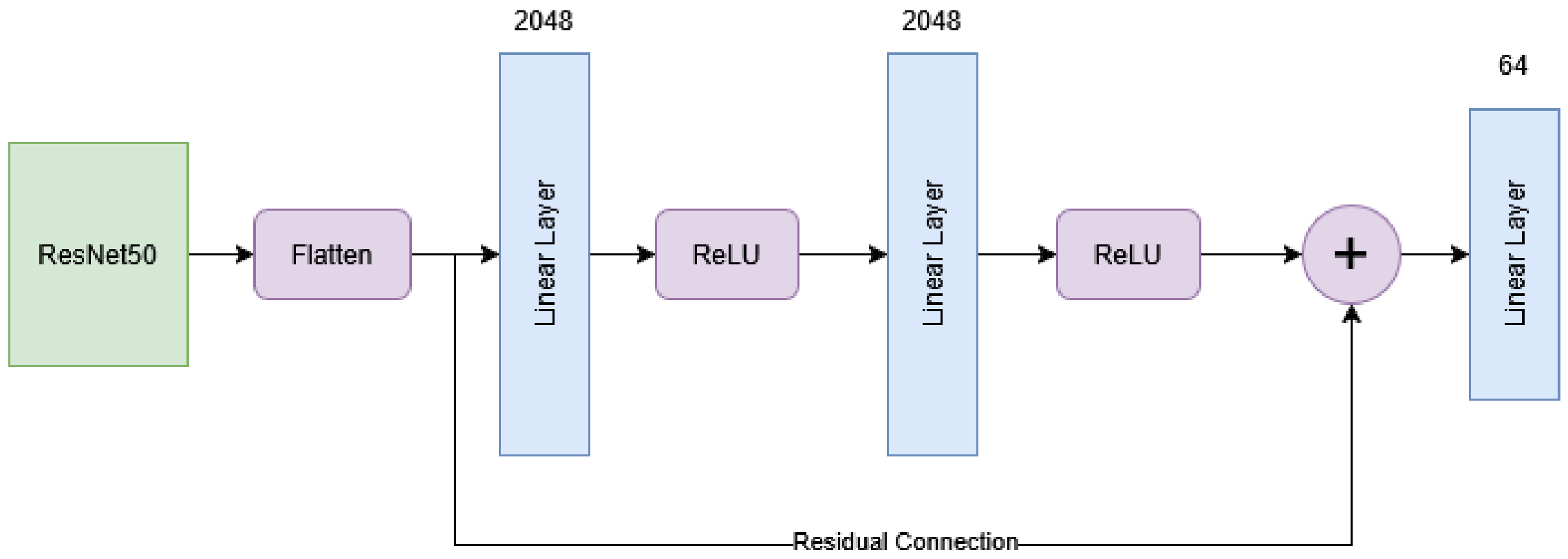

**Figure 10**: Architecture of word image embedding model. The head network has two first linear layers with an output size of 2048 and the last linear layer with an output size of 64, which is the chosen embedding size.

The model might be used in some scenarios if the word classifier is "not sure" about what word it sees, meaning the Softmax layer is not pointing to a specific class but rather spreading the probabilities among many classes. Our dataset does not cover the whole Latin vocabulary, so such cases might occur.

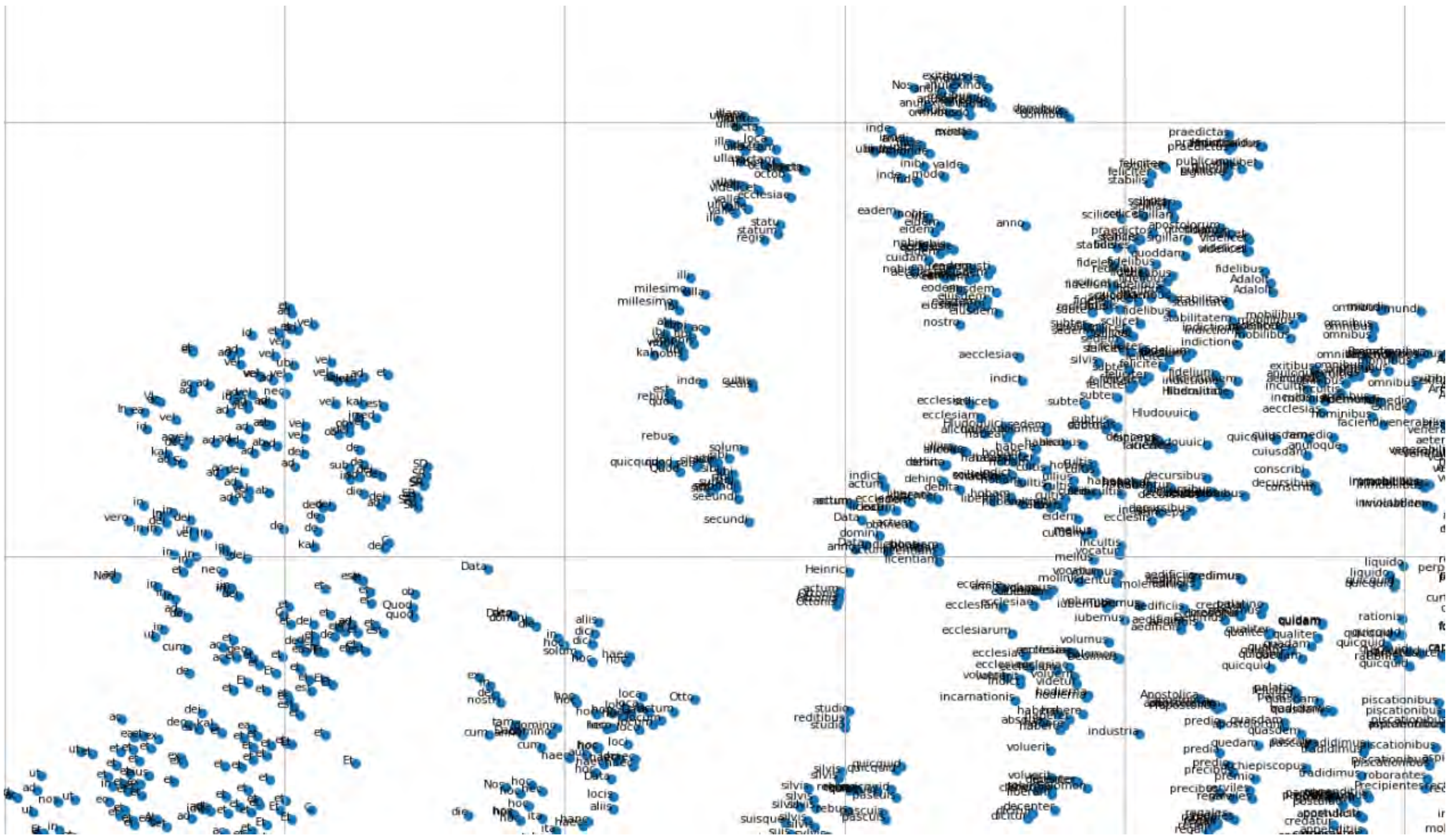

**Figure 11**: Part of the embedding space projected onto 2d space.

## 4.5. Full pipeline

The full pipeline is built using the models described in previous sections; the diagram of our pipeline is depicted in Figure 12. Firstly, we detect text lines using an image of the whole document. After cropping, the text lines are post-processed and sorted from top to bottom.

Each cropped line is processed by a word detection model to define the bounds of each word. Detected objects are post-processed, cropped, and sorted from left to right.

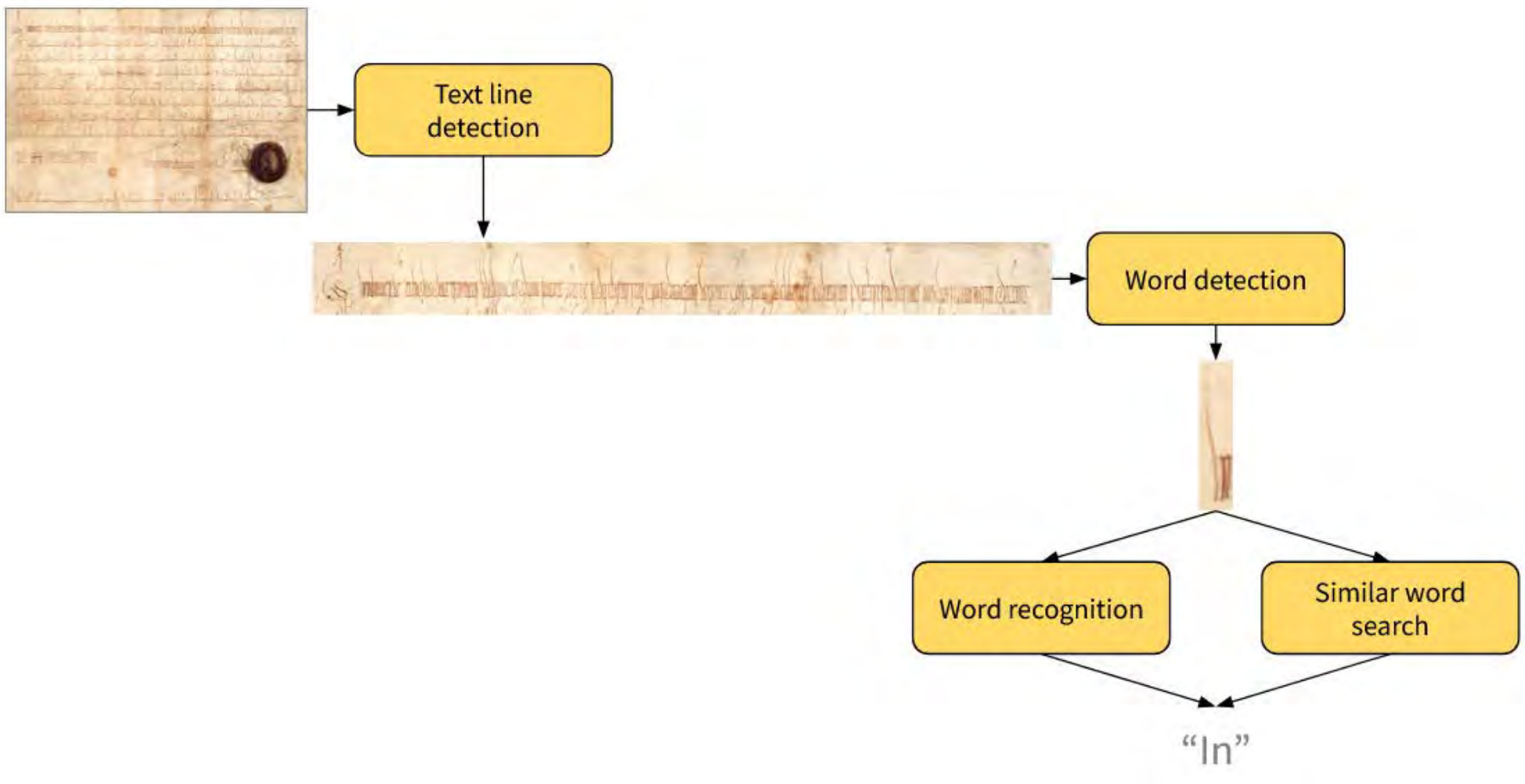

**Figure 12**: Transcription pipeline using detection, classification, and embedding models.

Each cropped word is classified to define its value. In case of too low confidence in classification, the pipeline returns a list of similar words, which are found by the embedding model. The word image is embedded by such a model, and the algorithm searches for the closest embeddings in a vector database.

The ordered classifications are returned as a list of predicted words. Because the main users of our solution are professional historians, the detected objects are also shown for more clarity and understanding of what's going on.

## 5. Results

Table 1 shows the results of our trained YOLO text line detection model. Figure 13 and Figure 14 show the Precision-Recall and F1 curves of text lines and words detection models, respectively. Table 2 and Table 3 show the confusion matrices of text lines and word detection models, respectively.

**Table 1**
YOLO detection metrics

| Model | Precision | Recall | mAP@50 | mAP@95 |
|---|---|---|---|---|
| Line detection | 0.98048 | 0.93388 | 0.97327 | 0.57584 |
| Words detection | 0.82679 | 0.70828 | 0.78447 | 0.42566 |

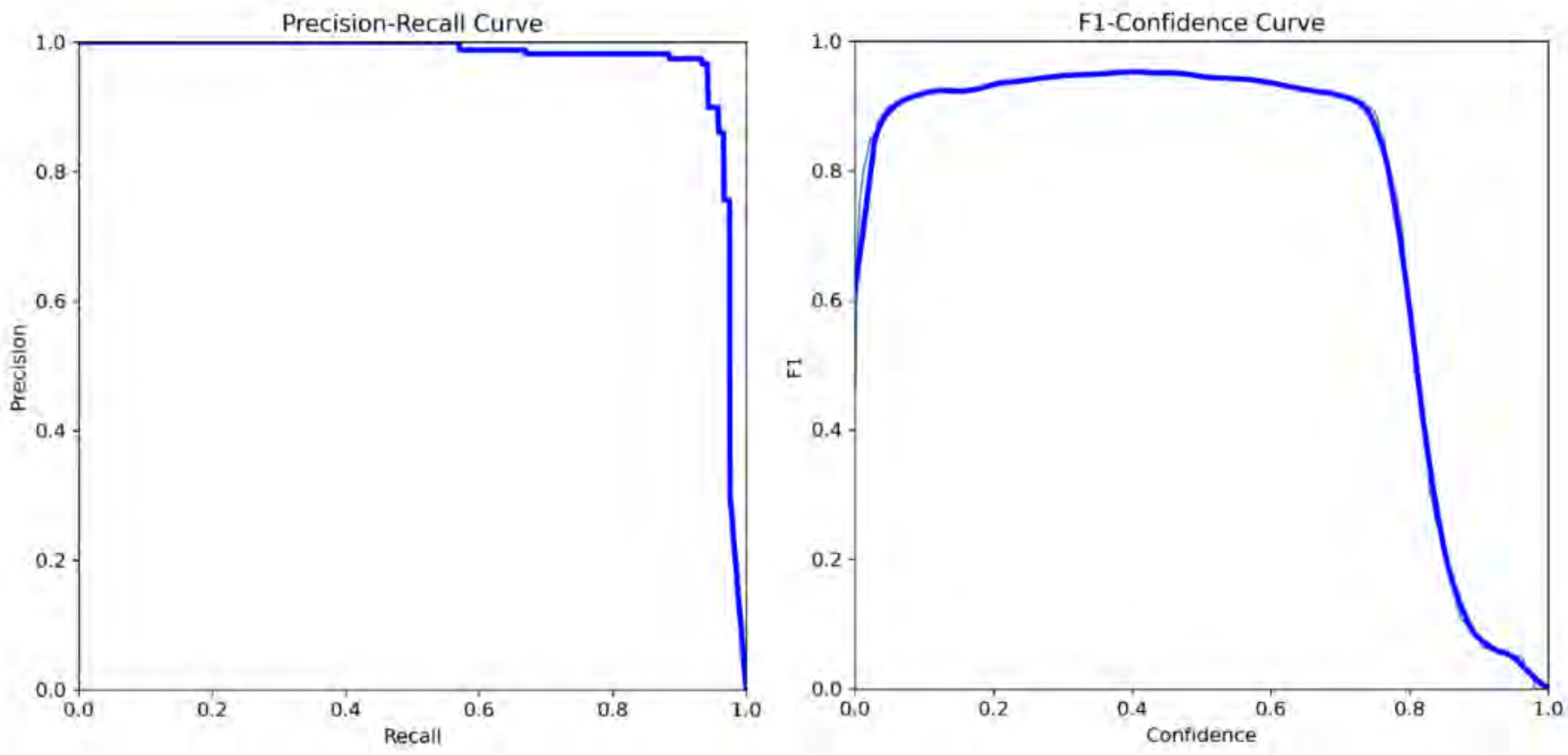

**Figure 13**: Precision-Recall Confidence and F1-Confidence curves of text line detection model.

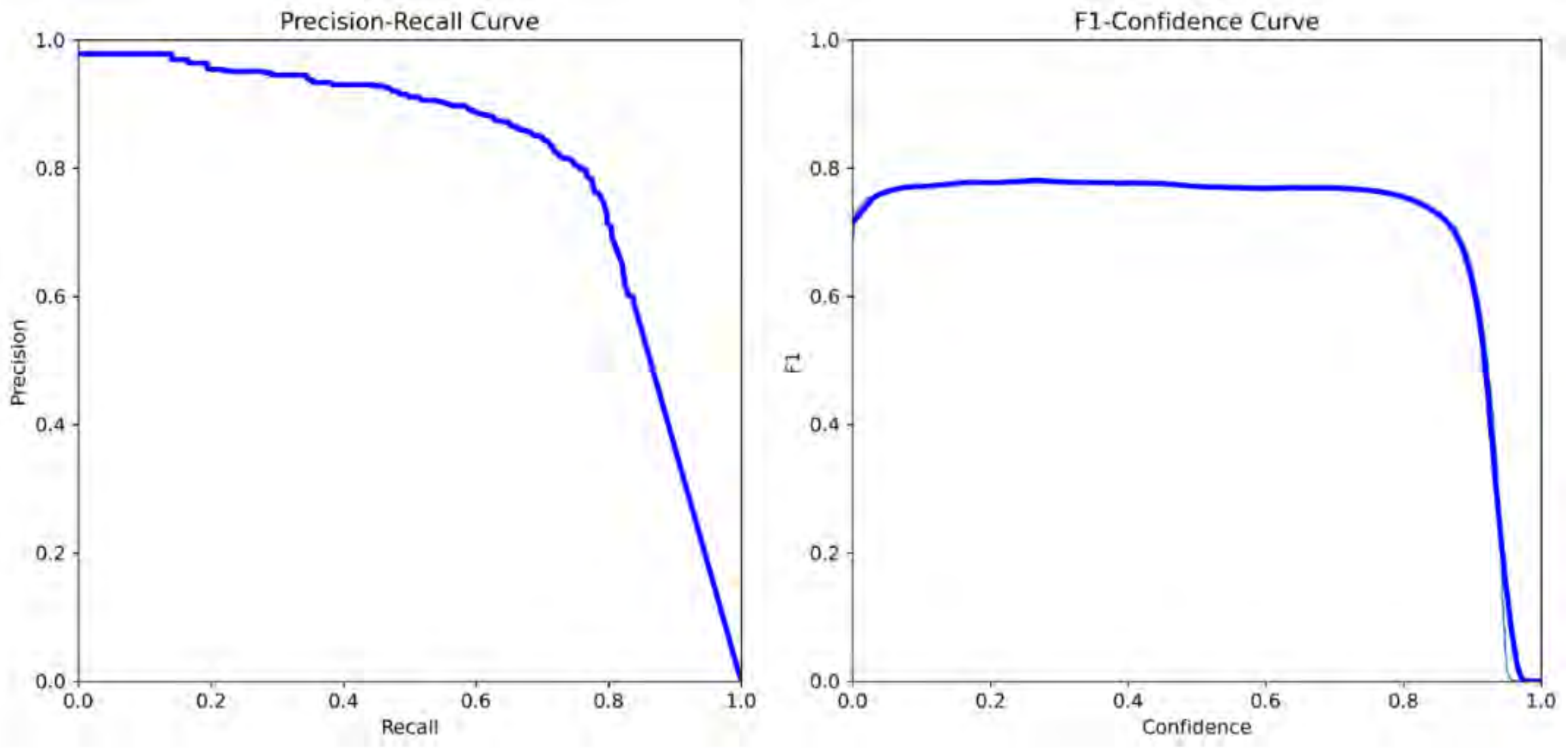

**Figure 14**: Precision-Recall Confidence and F1 Confidence curves of the word detection model.

**Table 2**
Text lines detection confusion matrix

| | Ground truth line | Ground truth background |
|---|---|---|
| Predicted line | 115 | 6 |
| Predicted background | 6 | - |

**Table 3**
Text lines detection confusion matrix

| | Ground truth word | Ground truth background |
|---|---|---|
| Predicted word | 750 | 170 |
| Predicted background | 207 | - |

To measure the results of the word image embedding model, we used a modified custom precision metric, which works in the following way: for the resulting embedding vector, find K nearest embeddings from a database and calculate the fraction of correct neighbors. Table 4 shows the connection between the number of nearest words and modified precision values.

**Table 4**
Modified precision values

| N closest words | Modified precision value |
|---|---|
| 1 | 0.5491 |
| 2 | 0.5027 |
| 3 | 0.5009 |
| 4 | 0.4772 |
| 5 | 0.4679 |
| 6 | 0.4545 |
| 7 | 0.4513 |
| 8 | 0.4431 |
| 9 | 0.4331 |

The example of word recognitions you can see on the Figure 15. In this example you can see the result of combining classification and word embedding models.

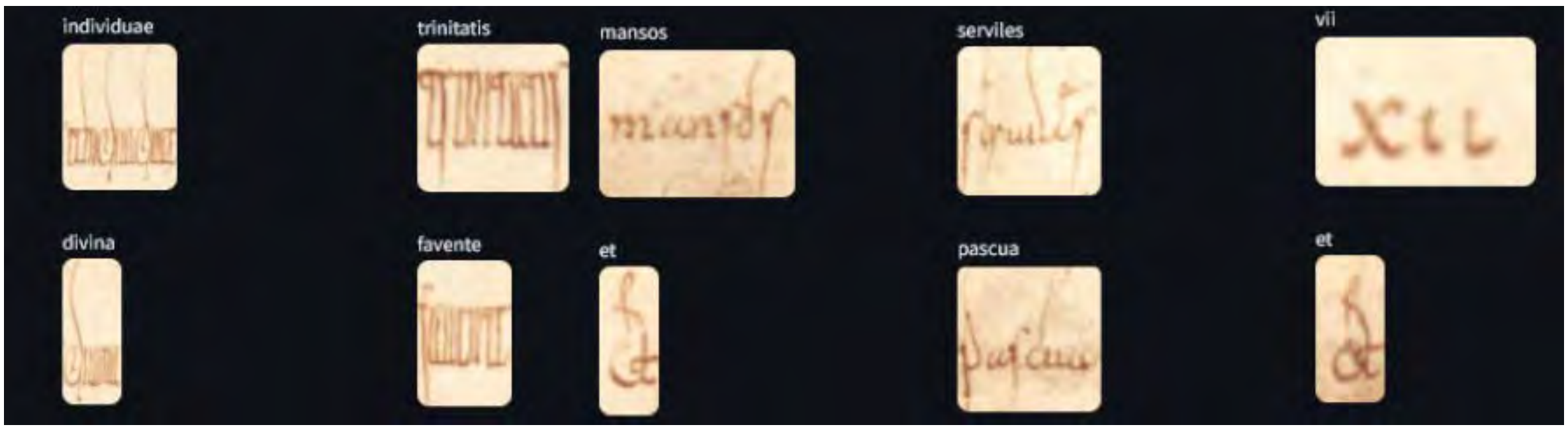

**Figure 15**: Example of recognized words using the combination of the classification and embedding models.

To measure the performance of our classification and embedding model in pairs, we used the mean string distance between predicted and ground truth words. The decision to use this metric was taken because of the high number of similar words in our data, so interpreting, for example, "nostri" and "nostro" as two different classes is not the best approach. On our data, the classification model was evaluated to a value of 2.4, meaning on average the predicted word differs from ground truth by 2-3 symbols.

# 6. Discussion

Our solution shows the effectiveness of such a combination of approaches in the case of transcribing text from handwritten medieval documents. The implementation of training and inference processes, explanatory data analysis, and dataset building utilities can be accessed on the GitHub repository [17].

## 6.1. Interpretation of results

Training of detection models has shown that such a relatively small amount of training data is enough to generalize it to be useful in unseen scenarios.

The embedding model maps the syntax information of a word using its image as input. 2D visualization of train embeddings shows the clusters of similar words. The vector databases allow for finding the closest vector representations fast enough for convenient usage. Metric results show the effectiveness of using the embedding model on naturally distributed data.

The classification model has shown good results, according to the nature of the input training data. It results in good results in the case of predicting common words and can optimize manual work for professional historians.

The whole pipeline consists of separate components; in other words, it's modular, which allows for upgrading the performance of the pipeline by affecting only components that are responsible for the features that need to be upgraded.

## 6.2. Comparison with previous research

In comparison with previous research, the main advantage of our approach is that it is more specific to medieval handwritings.

Our model is more applicable for professional usage because of its ability to see the transcription of each individual word, which helps to understand the whole text. This feature is not present in CTC-based solutions.

Text line detection has shown effective results in comparison to image segmentation approaches. It allows for correctly distinct, separate lines and easily crops them. The usage of oriented bounding boxes is a simple but also a fundamental decision, which makes the model capable of dealing with the natural style of handwritten text. Besides a relatively small amount of training data, text line detection has shown surprisingly good results.

Our pipeline is more adapted for processing documents in poorer condition, so the solution can be used for texts with stains, gaps, and other damage. Most of the public data consists of well-preserved document images, but the solution of this study shows methods to deal with damaged examples as well. For example, the system avoids recognizing damaged or lost parts of the document, because only the words that were detected are processed by the last 2 models of the pipeline.

The main disadvantage of our approach is that the pipeline models do not process context information. That might help understand written language, instead of separate words. Also, the system requires more manual work to be done to create the training datasets, especially detection ones. CTC approaches, where only image and text are required without annotated bounding boxes, there is no need to annotate each word.

### 6.3. Implications and limitations

Our work introduces a new approach to historical document processing, combining detection and recognition techniques. Even with a relatively small amount of data, the models have shown confident results. The solution can be improved or adapted by giving more annotated data.

Our classification dataset does not cover the whole Latin vocabulary, which may affect the recognition of documents with a large number of unseen words. Also, our models are trained on documents of a specific period and "genre", so the usability of our trained models does not apply to documents of a different style.

Our classification and embedding models perform much poorer on rare words; more annotated data is needed to improve this behavior.

Each model in our solution is dependent on the previous one in the pipeline, so if a line is not detected, we lose each word written in this line. Same for words, each undetected one cannot be classified and transcribed. Because of that, the error might sum up from the start to the end of the pipeline. Generative AI was not used for the study.

## 7. Conclusions

Our team has built an HTR solution suitable for extracting text information from handwritten medieval Latin documents. To deal with features specific to medieval documents, such as word contraction, damage, etc., we've come up with our own approach, based on a combination of detection, classification and embedding models.

We've annotated training datasets for detection, classification, and embedding models based on Latin documents from the $9^{th}$ to $11^{th}$ centuries. As a result, we have 31 images with annotated words and text lines. During the post processing we got a classification dataset, which maps word images to corresponding words.

We've trained text line and word detection models to locate and order objects we want to recognize. Post-processing steps were developed to transform detection data into the needed form. To deal with word contractions, we use word classification. Some words have a small number of occurrences in our data, so the classification model might struggle with predicting them correctly. Because of this, the embedding model was built, which encodes the cropped word image into an embedding space. To find the most similar words vector database is used.

The approach shows good results despite a relatively low amount of training data, which proves its efficiency in further use with larger datasets. There is a wide field of future improvements, such as context-aware text recognition and more detailed document structure analysis.

## Declaration on Generative AI

The authors have not employed any Generative AI tools.

## References

[1] Haus-, Hof- und Staatsarchiv, Salzburg, Erzstift (798-1806), in: Monasterium.net, URL https://www.monasterium.net/mom/AT-HHStA/SbgE, accessed 2025-03-22.

[2] Haus-, Hof- und Staatsarchiv, Salzburg, Domkapitel (831-1802), in: Monasterium.net, URL https://www.monasterium.net/mom/AT-HHStA/SbgDK/, accessed 2025-03-22.

[3] Phillip Benjamin Ströbel, Simon Clematide, Martin Volk. "Transformer-based HTR for Historical Documents", arXiv preprint, arXiv:2203.11008, 2022. doi: 10.48550/arXiv.2203.11008.

[4] Mohamed Yousef, Tom E. Bishop. "Origaminet: Weakly-Supervised, Segmentation-Free, One-Step, Full Page Text Recognition by learning to unfold", arXiv preprint, arXiv:2006.07491, 2020. doi: https://arxiv.org/abs/2006.07491.

[5] Christoph Wick, Jochen Zollner, Tobias Gruning. "Rescoring Sequence-to-Sequence Models for Text Line Recognition with CTC-Prefixes", arXiv preprint, arXiv:2110.05909, 2021. doi: https://arxiv.org/abs/2110.05909.

[6] Berat Barakat, Ahmad Droby, Majeed Kassis, Jihad El-Sana. "Text Line Segmentation for Challenging Handwritten Document Images Using Fully Convolutional Network", arXiv preprint, arXiv:2101.08299, 2021. doi: https://arxiv.org/abs/2101.08299.

[7] Ronneberger, O., Fischer, P., Brox, T. (2015). U-Net: Convolutional Networks for Biomedical Image Segmentation. In: Navab, N., Hornegger, J., Wells, W., Frangi, A. (eds) Medical Image Computing and Computer-Assisted Intervention – MICCAI 2015. MICCAI 2015. Lecture Notes in Computer Science(), vol 9351. Springer, Cham. https://doi.org/10.1007/978-3-319-24574-4_28.

[8] J. Redmon, S. Divvala, R. Girshick and A. Farhadi, "You Only Look Once: Unified, Real-Time Object Detection," 2016 IEEE Conference on Computer Vision and Pattern Recognition (CVPR), Las Vegas, NV, USA, 2016, pp. 779-788, doi: 10.1109/CVPR.2016.91.

[9] Büttner J, Martinetz J, El-Hajj H, Valleriani M. CorDeep and the Sacrobosco Dataset: Detection of Visual Elements in Historical Documents. J Imaging. 2022 Oct 15;8(10):285. doi: 10.3390/jimaging8100285. PMID: 36286379; PMCID: PMC9605005.

[10] Terven, Juan, Diana-Margarita Córdova-Esparza, and Julio-Alejandro Romero-González. 2023. "A Comprehensive Review of YOLO Architectures in Computer Vision: From YOLOv1 to YOLOv8 and YOLO-NAS" Machine Learning and Knowledge Extraction 5, no. 4: 1680-1716. https://doi.org/10.3390/make5040083.

[11] Rahal, N., Vögtlin, L. & Ingold, R. Historical document image analysis using controlled data for pre-training. IJDAR 26, 241–254 (2023). https://doi.org/10.1007/s10032-023-00437-8.

[12] M. Kozlenko, O. Zamikhovska, V. Tkachuk, L. Zamikhovskyi, Deep learning based fault detection of natural gas pumping unit, in: 2021 IEEE 12th International Conference on Electronics and Information Technologies (ELIT), Lviv, Ukraine, May 19-21, 2021, pp. 71-75, doi: 10.1109/ELIT53502.2021.9501066.

[13] M. Kozlenko, V. Sendetskyi, O. Simkiv, N. Savchenko, A. Bosyi, Identity documents recognition and detection using semantic segmentation with convolutional neural network, in: 2021 Workshop on Cybersecurity Providing in Information and Telecommunication Systems, CEUR Workshop Proceedings, vol. 2923, Kyiv, Ukraine, Jan. 28, 2021, pp. 234-242. https://ceur-ws.org/Vol-2923/paper25.pdf.

[14] R. W. Hamming, "Error detecting and error correcting codes," in The Bell System Technical Journal, vol. 29, no. 2, pp. 147-160, April 1950, doi: 10.1002/j.1538-7305.1950.tb00463.x.

[15] Zhuang, Fuzhen et al. "A Comprehensive Survey on Transfer Learning." Proceedings of the IEEE 109 (2019): 43-76.

[16] Peng, Peng & Wang, Jiugen. (2020). How to fine-tune deep neural networks in few-shot learning?. 10.48550/arXiv.2012.00204.

[17] M. Voloshchuk, B. Zarembovska, Carolingus Project, 2025. URL: https://github.com/AIVMZB/Carolingus.